\documentclass[letterpaper]{article} 
\usepackage{aaai2026}  
\usepackage{times}  
\usepackage{helvet}  
\usepackage{courier}  
\usepackage[hyphens]{url}  
\usepackage{graphicx} 
\usepackage{natbib}  
\usepackage{caption} 
\usepackage{amsmath}

\usepackage{algorithm}
\usepackage{algorithmic}
\usepackage{amssymb}  

\usepackage{newfloat}
\usepackage{listings}
\DeclareCaptionStyle{ruled}{labelfont=normalfont,labelsep=colon,strut=off} 
\floatstyle{ruled}
\newfloat{listing}{tb}{lst}{}
\floatname{listing}{Listing}

\usepackage{booktabs}
\usepackage{multirow}

\usepackage{makecell}
\usepackage{pifont}
\usepackage{subcaption}  

\title{Understanding Syllogistic Reasoning in LLMs from Formal and Natural Language Perspectives}

\author{%
Aheli Poddar\textsuperscript{\rm 1},
Saptarshi Sahoo\textsuperscript{\rm 2},
Sujata Ghosh\textsuperscript{\rm 2}
}

\affiliations{%
\textsuperscript{\rm 1}Institute of Engineering \& Management, Kolkata \\
\textsuperscript{\rm 2}Indian Statistical Institute, Chennai \\
aheli.poddar2022@iem.edu.in, saptarshi@isichennai.res.in, sujata@isichennai.res.in
}

\begin{document}
\maketitle

\begin{abstract}
We study syllogistic reasoning in LLMs from the logical and natural language perspectives. In process, we explore  fundamental reasoning capabilities of the LLMs and the direction this research is moving forward. To aid in our studies, we use 14 large language models and investigate their syllogistic reasoning capabilities in terms of symbolic inferences as well as natural language understanding. Even though this reasoning mechanism is not a uniform emergent property across LLMs, the perfect symbolic performances in certain models make us wonder whether LLMs are becoming more and more formal reasoning mechanisms, rather than making explicit the nuances of human reasoning.
\end{abstract}



\begin{links}
    \link{Code}{https://github.com/XAheli/Logic-in-LLMs}
\end{links}


\section{Introduction}

With the unprecedented development of large language models (LLMs) in recent years that have made them resemble human speakers and reasoners to a great extent in many levels \cite{holliday2024conditional,bubeck2023ece,zhao2023survey}, the reasoning capabilities of LLMs have increased manifold. To motivate such growth, 
the question we generally ask an LLM is to what extent the LLM has grasped logical reasoning in its different forms, for example, see \cite{holliday2024conditional,borazjanizadeh2024reliable,sambrotta2025llms}. In contrast, the motivation for this study is somewhat distinct in nature in that we wonder whether developing LLM to have excellent logical reasoning capabilities is fruitful in the long run, as having such features does not bring an LLM closer to mimicking human reasoning. As a case in point, we consider syllogistic reasoning from a formal as well as natural language viewpoint. 

Evidently, humans are far from logical when it comes to reasoning, and they are often influenced by their past experiences and knowledge, for example, 
consider the belief-bias effect \cite{evans1983conflict}: People doing syllogistic reasoning are often influenced by the believability of the conclusion. In fact, 
it is shown by \cite{lewton2016relationship} that individuals with autistic traits show less belief-bias effect than typical individuals. In this scenario, one might consider to check whether LLM reasoning is close to human reasoning by studying the belief-bias effect on the LLMs, and the present work studies this question. We note that \cite{eisape2024systematic} studied a similar question, but their methodology is quite different from ours. Before describing the exact contribution of this work, let us discuss some recent work on syllogistic reasoning in LLMs.

A novel framework dealing with legal syllogistic reasoning is provided in \cite{zhang2025syler}. In this work, the LLMs are empowered to provide explicit and trustworthy legal reasoning by integrating a retrieval mechanism with reinforcement learning. A mechanistic interpretation of syllogistic reasoning is provided in \cite{kim2025reasoning}. This work deals with belief-biases as well and it is shown that such biases contaminate the reasoning mechanisms. In \cite{zong2024categorical}, the authors make a detailed survey on the reasoning capabilities of LLMs with respect to categorical syllogisms. 


This work makes several key contributions to understanding syllogistic reasoning in LLMs from both formal and natural language perspectives. We introduce a novel dual ground truth framework that evaluates each syllogism on two separate dimensions: syntactic validity \textit{(does the conclusion logically follow?)} and natural language believability \textit{(is the conclusion intuitively plausible?)}. These two dimensions may align or conflict with each other, enabling us to assess formal reasoning capabilities independently from natural language understanding. Through a comprehensive empirical study, we systematically evaluated 14 state-of-the-art LLMs across four prompting strategies and three temperature settings on carefully constructed syllogisms covering diverse logical structures and belief-bias conditions. Our analysis reveals that the majority of models exhibit a significant measure of belief bias; in other words, they perform better on certain kinds of problem (where logic aligns with intuition) than others. We further uncover a substantial gap between syntactic and natural language understanding accuracy, demonstrating that current LLMs excel at formal logical structure while struggling with natural language plausibility judgments---a pattern opposite to human reasoning tendencies. Contrary to conventional wisdom, we find that few-shot prompting degrades performance compared to zero-shot, and that reasoning capability depends critically on architectural choices rather than raw parameter count. These findings raise a fundamental question: Are LLMs evolving into formal reasoning engines that surpass human-like reasoning with its inherent biases?

The remainder of the paper is structured as follows. Section~\S\ref{sec:background} provides a brief overview of syllogisms. Section~\S\ref{sec:experiments} delves into the experimental details, including the models, data, overall methodology, prompting variants, and evaluation metrics. Section~\S\ref{sec:results} reports on the findings and their interpretations. Section~\S\ref{sec:limitations} provides a discussion of the limitations of our study, and Section~\S\ref{sec:futurework} concludes the article.



\section{On Syllogisms}
\label{sec:background}

The concept of \textit{syllogism} was first introduced by Aristotle \cite{smith1989prior}, and as observed by Robin Smith \cite{smith2017logic}, a syllogism in modern logic consists of three subject-predicate propositions, two premises, and a conclusion, and whether or not the conclusion follows from the premises. An example of syllogism is as follows: \textit{``No footballer is a swimmer; Some swimmers are gardeners; Therefore, some gardeners are not footballers.''} When terms like \emph{footballer} or \emph{swimmer} are replaced by generic terms like B, C and D, we can rewrite the above premises by: \textit{``No B is C; Some C are D.''}  A conclusion relates the non-shared terms, for example, \textit{``Some D are not B''}. 

In the literature, various types of syllogisms are studied, categorical, conditional, and others \cite{copi2016introduction}. In this work, we mostly concentrate on categorical syllogisms, but we consider a few others as well. The statements of a categorical syllogism look like the following: \textit{Quantifier (Subject) Copula (Predicate),} which take four standard forms, viz.
	
\begin{itemize}
\item[-] \textit{Universal Affirmative (A)}: All S are P, i.e., $S \subseteq P$.
\item[-] \textit{Universal Negative (E)}: No S is P, i.e., $S \cap P = \emptyset$.
\item[-] \textit{Particular Affirmative (I)}: Some S is P, i.e., $S \cap P \ne \emptyset$.
\item[-] \textit{Particular Negative (O)}: Some S is not P, i.e., $S \setminus P \ne \emptyset$.
\end{itemize}

Here, S is the subject and P is the predicate. S and P are generally termed variables, and these quantifier styles, namely, A, E, I, O, are called `moods'. The variables may change their orders, leading to new premises. As mentioned earlier, one of the three variables used in a syllogism is not there in the conclusion, and evidently the variable is common to both premises. Depending on the placement of the common variable (C, say) that does not occur in the conclusion, we get four types of \textit{figures} for syllogisms. See Table~\ref{tab:figure} for a detailed description.
       
\begin{table}[t]
\centering
\begin{tabular}{cccc}
\toprule
\textbf{1} & \textbf{2} & \textbf{3} & \textbf{4} \\
\midrule
B-C & C-B & B-C & C-B \\
C-D & D-C & D-C & C-D \\
\bottomrule
\end{tabular}
\caption{A description of the four figures for syllogisms containing the variables B, C, and D.}
\label{tab:figure}
\end{table}

We should note here that, in statements of type A, `All' is sometimes overlooked for the sake of simplicity. The following example clarifies the point: \textit{``All vehicles have wheels; Boats are vehicles / A boat is a vehicle; Therefore, boats have wheels / a boat has wheels.''}

A syllogism is said to be {\it valid} if the truth of the premises implies the truth of the conclusion.
A way to check the validity of a syllogism is by converting the statements in a suitable first order language and check the validity there. The other way is through enumerating each case (there will be some finite number of cases where the two premises will have one of the four forms A, E, I or O) and then using standard Venn Diagram techniques to fix the conclusion. Thus, when a new tuple of syllogism comes in, the job of checking validity boils down to just checking the instance from the already defined cases and to conclude from it. 

A syllogism is said to be {\it believable} if the conclusion of the syllogism is actually true. For this case, the logical argument does not play any role. The main goal of this research work is two-fold. On one hand, we would like to check how accurately the LLMs can do syllogistic reasoning, and on the other hand we would like to check whether context and real world knowledge play any role in their reasoning processes. To this end, the following four categories of syllogisms play a significant role, namely (i) valid-believable, (ii) valid-unbelievable, (iii) invalid-believable, and (iv) invalid-unbelievable. These distinct types are summarized in Table~\ref{tab:syllogismTypes}, given in \cite{brauner2025understanding}, which provides an example for each such type of syllogism. 

\begin{table*}[t]
\centering
\begin{tabular}{l|l|l} 
\toprule
& \textbf{Believable} & \textbf{Unbelievable} \\ 
\midrule
\textbf{Valid} & \textit{All birds have feathers} & \textit{All mammals walk} \\
& \textit{Robins are birds} & \textit{Whales are mammals} \\ 
& \textit{Therefore robins have feathers} & \textit{Therefore whales walk} \\
\midrule
\textbf{Invalid} & \textit{All flowers need water} & \textit{All insects need oxygen} \\
& \textit{Roses need water} & \textit{Mice need oxygen} \\ 
& \textit{Therefore roses are flowers} & \textit{Therefore mice are insects} \\
\bottomrule
\end{tabular}
\caption{Example syllogisms illustrating the four categories described in \S\ref{sec:background}.}
\label{tab:syllogismTypes}
\end{table*}

    



\section{Experiments}
\label{sec:experiments}
We conduct a systematic evaluation of syllogistic reasoning capabilities across diverse language models, examining the effects of prompting strategies, temperature settings, and content variations on logical inference accuracy. Our experimental design encompasses 168 unique configurations (14 models $\times$ 4 strategies $\times$ 3 temperatures), enabling comprehensive analysis of factors influencing LLM syllogistic reasoning performance. 

\subsection{Models}



We evaluated syllogistic reasoning capabilities in 14 large language models spanning 8 organizations, listed in Table~\ref{tab:overall_performance}. The Google Gemini models were accessed through Google AI Studio APIs.\footnote{\texttt{https://ai.google.dev/gemini-api/docs}} All remaining models were accessed via the HuggingFace Inference API\footnote{\texttt{https://huggingface.co/docs}} using the \textit{:cheapest} routing for automatic provider selection.\footnote{Total API costs for all experiments were approximately \$ 500} Our model selection prioritized four criteria: (1) organizational diversity to capture different development philosophies, (2) parameter scale range (1B to 671B) to assess scaling effects, (3) architectural variety including dense transformers and Mixture-of-Experts (MoE) systems, and (4) API reproducibility. 


\subsection{Data and Methodology}

\subsubsection{Dataset Construction}

For our experiments, we constructed a benchmark of 160 syllogisms, mostly categorical, adapted from the cognitive science and psychology literature on human syllogistic reasoning \cite{solcz2008role,lewton2016relationship}.

We began with 40 base syllogisms, each handcrafted to cover different syllogistic figures and validity conditions. To isolate the effects of logical structure from natural languagecontent, given our \textit{dual ground truth annotations}, we created three additional variants for each base syllogism. The \textit{nonsense variant} (X) replaces meaningful predicates with abstract terms (e.g., \textit{``blargs''}, \textit{``zimons''}, \textit{``glorps''}), testing pure logical reasoning without natural language interference. The \textit{order-switched variant} (O) reverses the order of presentation of the premises to test the sensitivity to the structure of the argument. The \textit{combined variant} (OX) applies both modifications, providing a comprehensive robustness assessment.

For example, the normal variant \textit{``All calculators are machines; All computers are calculators; Therefore, some machines are not computers''} becomes \textit{``All blargs are zimons; All glorps are blargs; Therefore, some zimons are not glorps''} in its nonsense form. We reviewed all stimuli and made necessary adjustments by hand to ensure grammatical correctness and logical equivalence across variants.

\subsubsection{Dual Ground Truth}
\label{subsec:dualgroundtruth}

\begin{figure}[t]
\centering
\includegraphics[width=0.9\columnwidth]{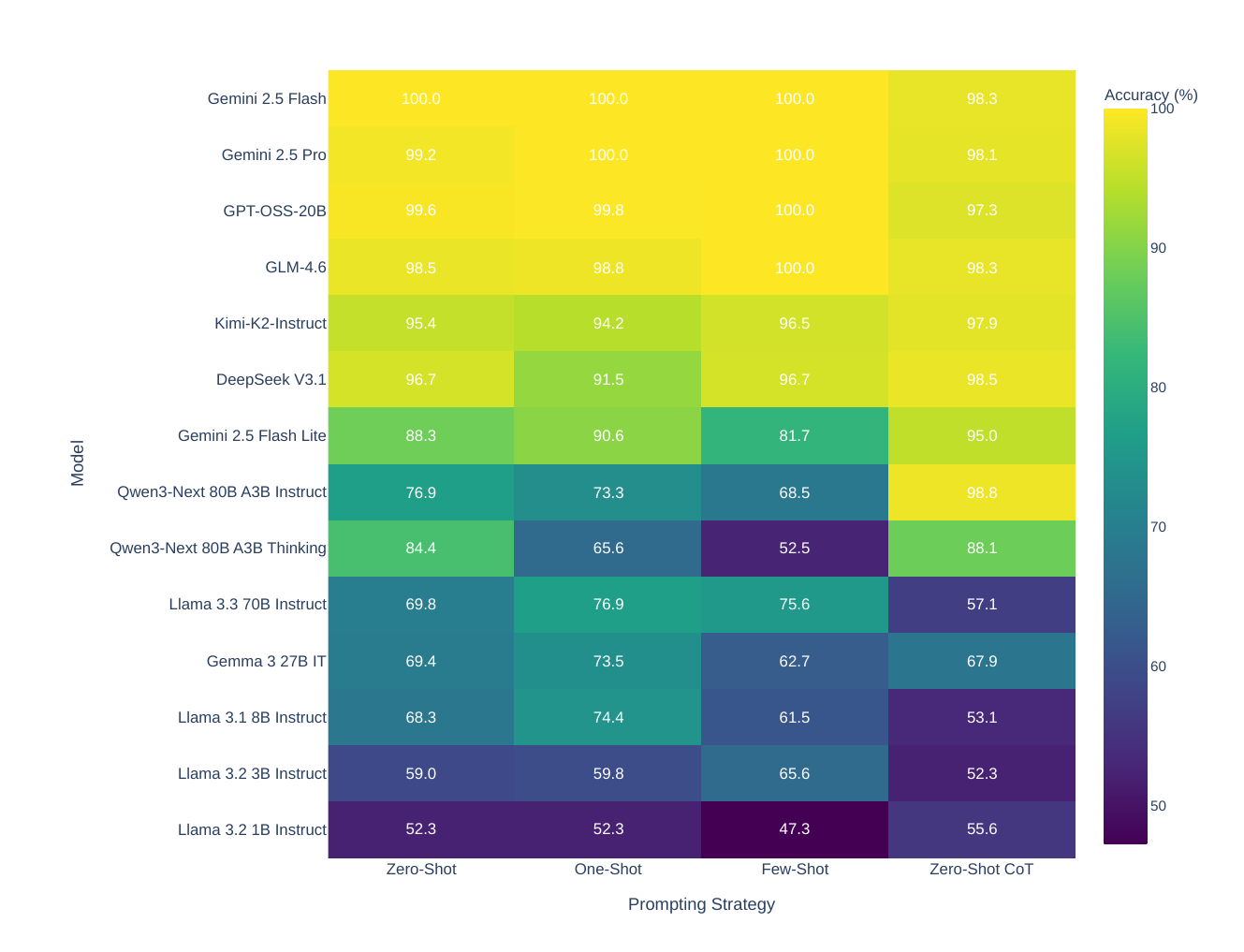}
\caption{Heatmap of model accuracy across four prompting strategies (Zero-shot, One-shot, Few-shot, Zero-shot Chain-of-Thought). Despite few-shot showing significant mean decline ($\Delta=-3.57$ pp, $p=0.0165^*$), systematic patterns across models remain minimal, indicating strategy effects are model-specific rather than universal.}
\label{fig:strategy_heatmap}
\end{figure}

Each syllogism carries two independent ground truth annotations, enabling orthogonal evaluation of logical reasoning and natural language processing. The \textit{syntactic validity label} (valid/invalid) indicates whether the conclusion logically follows from the premises according to formal syllogistic rules, independent of content truth. The \textit{natural language understanding (NLU) label} (believable/unbelievable) indicates whether the conclusion is intuitively plausible given real-world knowledge, independent of logical structure.

The dataset comprises 76 valid syllogisms (47.5\%) and 84 invalid syllogisms (52.5\%). For believability, 38 instances (23.8\%) have believable conclusions while 122 (76.2\%) have unbelievable or abstract conclusions. This asymmetry reflects the inclusion of nonsense variants, which by design have semantically neutral conclusions.

\begin{figure}[t]
\centering
\includegraphics[width=0.9\columnwidth]{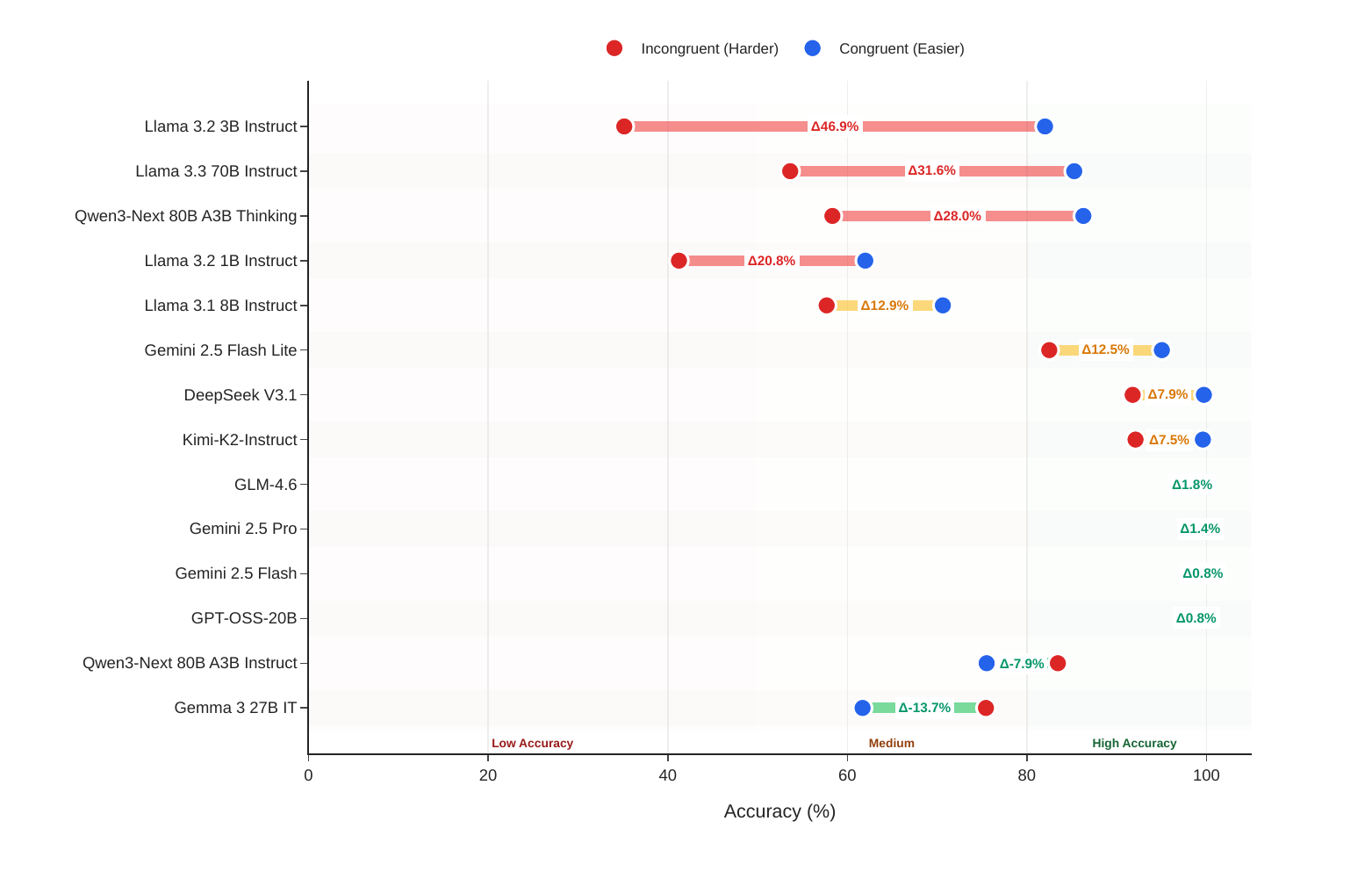}
\caption{Belief bias effect across 14 models comparing performance on congruent syllogisms (logic aligns with intuition) versus incongruent syllogisms (logic conflicts with intuition). Twelve models (86\%) exhibit positive bias ($\Delta=+10.81$ pp, $p=0.0280^*$, $d=0.66$). Top-tier models show minimal bias ($<2$ pp), while lower-tier models show severe bias (up to $+46.9$ pp). Negative correlation ($\rho=-0.565^*$) indicates higher reasoning ability reduces reliance on semantic heuristics.}
\label{fig:belief_bias}
\end{figure}







\subsubsection{Belief Bias Categories}

Belief bias is a well-documented phenomenon in human cognition whereby 
reasoners accept logically invalid conclusions that seem plausible, or 
reject valid conclusions that seem implausible---allowing the semantic 
content of conclusions to override evaluation of logical structure 
\cite{evans1983conflict,klauer2000belief,pennycook2013belief}. 

Our dual annotation scheme enables formal quantification of this 
effect by categorizing syllogisms based on alignment between logical 
validity and intuitive believability:

\textbf{\textit{Congruent instances}} (82 instances, 51.2\%) are cases 
where logic and intuition align: valid-believable or invalid-unbelievable 
conclusions. These represent \textit{``easy''} cases where correct logical 
judgment matches intuitive response. 

\textbf{\textit{Incongruent instances}} (78 instances, 48.8\%) are cases 
where logic and intuition conflict: valid-unbelievable or invalid-believable 
conclusions. These \textit{``hard''} cases directly test whether models can 
override semantic plausibility with formal reasoning.

For example: \textit{``All things with an engine need oil; Cars need oil; 
Therefore, cars have engines.''} This conclusion is factually correct yet 
logically invalid (affirming the consequent fallacy). Such instances are 
particularly diagnostic, as accepting them indicates susceptibility to 
belief bias.


\subsection{Prompting Schema}

We implement four prompting strategies to evaluate models under varying levels of task specification and reasoning scaffolding: \textit{\textbf{Zero Shot (ZS)}} and \textit{\textbf{One-shot (OS)}}, which utilize zero and one demonstration example respectively to test intrinsic capability; \textit{\textbf{Few Shot (FS)}}, which provides four balanced examples (2 valid, 2 invalid) including a belief bias trap to distinguish natural language plausibility from logical validity; and \textit{\textbf{ZS Chain-of-Thought (ZS CoT)}}, which encourages intermediate reasoning traces \citep{kojima2022large}. Critically, regardless of the context or scaffolding provided, all strategies request the same final response format: a single word \textit{``correct''} or \textit{``incorrect''} to ensure comparability across conditions.

Algorithm~\ref{alg:syllogism} presents our unified inference procedure that adapts its behavior based on the temperature parameter $\tau$. The algorithm accepts a syllogism $\mathcal{S}$ consisting of two premises $p_1, p_2$ and a conclusion $c$, a prompting strategy $\sigma$, and outputs a validity prediction $\hat{y}$ along with a confidence score $\rho$.

\subsubsection{Strategy Specifications}

The procedure begins by constructing task-specific prompts through two subroutines. \textsc{BuildSystemPrompt}($\sigma$) generates the system-level instruction that defines the reasoning task:

\begin{center}
\fbox{\parbox{0.9\columnwidth}{%
\small
``You are an expert in syllogistic reasoning. Your task is to determine whether the conclusion of a given syllogism follows from the premises. A syllogism is CORRECT if the conclusion follows from the premises. A syllogism is INCORRECT if the conclusion does not follow. \textbf{\textit{[Strategy-specific addition.]}} Respond with exactly one word: `correct' or `incorrect'.''
}}
\end{center}

For ZS CoT, the system prompt appends \textit{``Think through step by step''} before the response instruction; all other strategies use identical system prompts. 
\textsc{BuildUserPrompt}($\mathcal{S}, \sigma$) constructs the user message by optionally including demonstration examples (1 for one-shot, 4 for FS), formatting the input syllogism with labeled premises and conclusion, and appending the query.

\begin{algorithm}[t]
\caption{Temperature-Adaptive Syllogistic Reasoning}
\label{alg:syllogism}
\begin{algorithmic}[1]
\REQUIRE Syllogism $\mathcal{S} = (p_1, p_2, c)$; Strategy $\sigma \in \{\text{ZS}, \text{OS}, \text{FS}, \text{ZSCoT}\}$; Temperature $\tau \in \{0.0, 0.5, 1.0\}$
\ENSURE Prediction $\hat{y} \in \{\text{valid}, \text{invalid}\}$; Confidence $\rho \in [0, 1]$
\STATE \textbf{Parameters:} $K_{\max} = 10$, $\eta = 5$ \COMMENT{Max samples, early stopping threshold}
\STATE
\STATE $\pi_{\text{sys}} \gets \textsc{BuildSystemPrompt}(\sigma)$
\STATE $\pi_{\text{user}} \gets \textsc{BuildUserPrompt}(\mathcal{S}, \sigma)$
\IF{$\tau = 0$}
    \RETURN $\textsc{Parse}(\textsc{Query}(\pi_{\text{sys}}, \pi_{\text{user}}, 0)), 1.0$
\ENDIF
\STATE $n_+ \gets 0$, $n_- \gets 0$
\FOR{$k = 1$ to $K_{\max}$}
    \STATE $\hat{y}_k \gets \textsc{Parse}(\textsc{Query}(\pi_{\text{sys}}, \pi_{\text{user}}, \tau))$
    \STATE $n_+ \gets n_+ + \mathbb{1}[\hat{y}_k = \text{valid}]$
    \STATE $n_- \gets n_- + \mathbb{1}[\hat{y}_k = \text{invalid}]$
    \IF{$k = \eta$ and $\min(n_+, n_-) = 0$}
        \STATE \textbf{break} \COMMENT{Early stop if unanimous}
    \ENDIF
\ENDFOR
\STATE $\hat{y} \gets \begin{cases} 
\text{valid} & \text{if } n_+ > n_- \\
\text{invalid} & \text{otherwise}
\end{cases}$ \COMMENT{Ties default to invalid}
\STATE $\rho \gets \max(n_+, n_-) / (n_+ + n_-)$
\RETURN $\hat{y}, \rho$
\end{algorithmic}
\end{algorithm}



\subsubsection{Adaptive Stopping Strategy}

When $\tau = 0$, the algorithm performs greedy deterministic decoding, querying the language model once, and returning the parsed prediction with full confidence ($\rho = 1.0$). For stochastic sampling ($\tau > 0$), we implement self-consistency \citep{wang2023selfconsistency} by generating up to $K_{\max} = 10$ independent samples. For each sample $k$, we query the model with temperature $\tau$ and parse the response to extract the validity label $\hat{y}_k$. We maintain counters $n_+$ and $n_-$ for valid and invalid predictions, respectively, using indicator functions $\mathbb{1}[\cdot]$.

To improve efficiency, we employ early stopping inspired by \citet{holliday2024conditional}: if the first $\eta = 5$ samples are unanimous (i.e., $\min(n_+, n_-) = 0$ at $k = \eta$), we terminate sampling. This reduces API calls substantially when models exhibit high confidence. The final prediction $\hat{y}$ is determined by majority vote. Any ties by default maps to \textit{``invalid''} as a conservative choice.


\subsection{Evaluation Methods}

\subsubsection{Primary Metrics}

We evaluate model responses using standard classification metrics: accuracy $(TP + TN) / N$, precision $TP / (TP + FP)$, recall $TP / (TP + FN)$, and F1 score as the harmonic mean of precision and recall. Accuracy serves as the primary metric given the near-balanced class distribution (47.5\% valid, 52.5\% invalid).

\subsubsection{Dual Evaluation Framework}

Each model prediction is evaluated against both ground truths independently. For syntactic evaluation, the model response maps ``correct'' $\to$ \textit{valid} and ``incorrect'' $\to$ \textit{invalid}, compared against \texttt{ground\_truth\_syntax}. For NLU evaluation, it maps ``correct'' $\to$ \textit{believable} and ``incorrect'' $\to$ \textit{unbelievable}, compared against \texttt{ground\_truth\_NLU}. This dual evaluation reveals whether models assess logical structure, natural language content, or some combination thereof.




\subsubsection{Belief Bias Effect}

Classical belief bias research employed indices derived from raw endorsement rates
\cite{evans1983conflict,klauer2000belief}. However, these traditional indices have been
criticized on psychometric grounds \cite{dube2010traditional,heit2014traditional}:
changes in proportions starting from different baseline values are not readily
comparable, and empirical receiver operating characteristic (ROC) curves reveal
curvilinear relationships that violate the linear assumptions of difference scores.

We adopt a direct accuracy-based approach aligned with recent studies 
\cite{trippas2014fluency}, quantifying belief bias as the accuracy differential 
between congruent and incongruent syllogisms:
\[
\Delta_{\text{bias}} = \text{Acc}_{\text{congruent}} - \text{Acc}_{\text{incongruent}}
\]
where ${Acc}_{{congruent}}$ is accuracy on valid-believable plus invalid-unbelievable 
instances (where logic and intuition align), and ${Acc}_{{incongruent}}$ is accuracy 
on valid-unbelievable plus invalid-believable instances (where they conflict).

This metric is appropriate for our setting because: (1) 
our LLM evaluations produce binary correct/incorrect judgments rather than 
confidence-rated responses, eliminating the ROC curvature concerns that motivated 
signal detection approaches \cite{dube2010traditional}; (2) accuracy percentages are directly interpretable and comparable across 
all conditions, unlike endorsement-rate indices which suffer from 
baseline-dependency \cite{heit2014traditional}; 
(3) our within-subjects design compares each model against itself on 
congruent versus incongruent trials, isolating the belief bias effect 
while controlling for differences in overall reasoning ability. 
Positive $\Delta_{\text{bias}}$ indicates susceptibility to belief bias i.e., the model performs better when semantic content aligns with logical structure than when they conflict.


\subsubsection{Consistency Metric}

We measure response consistency across content variants of logically equivalent syllogisms. Let $\mathcal{S}$ denote the set of 40 base natural syllogisms and $\hat{y}_{s,v}$ the model's prediction for syllogism $s$ under variant $v \in \{N, X, O, OX\}$. We define:
\begin{align}
C_{\text{all}} &= \frac{1}{|\mathcal{S}|} \sum_{s \in \mathcal{S}} \mathbb{1}\left[ \hat{y}_{s,N} = \hat{y}_{s,X} = \hat{y}_{s,O} = \hat{y}_{s,OX} \right] \\
C_{N \leftrightarrow X} &= \frac{1}{|\mathcal{S}|} \sum_{s \in \mathcal{S}} \mathbb{1}\left[ \hat{y}_{s,N} = \hat{y}_{s,X} \right] \\
C_{O \leftrightarrow OX} &= \frac{1}{|\mathcal{S}|} \sum_{s \in \mathcal{S}} \mathbb{1}\left[ \hat{y}_{s,O} = \hat{y}_{s,OX} \right]
\end{align}
where $C_{\text{all}}$ denotes overall consistency across all four variants. The pairwise metrics isolate specific invariance properties: $C_{N \leftrightarrow X}$ tests robustness to natural language content (meaningful vs. nonsense predicates), while $C_{O \leftrightarrow OX}$ tests robustness to premise order within matched content types.



\section{Results}
\label{sec:results}
Our evaluation comprises 26,880 model-instance evaluations (14 models $\times$ 4 strategies $\times$ 3 temperatures $\times$ 160 syllogisms). We report syntactic accuracy as the primary metric, with supplementary analyses of dual-framework evaluation, belief bias, variant robustness, and response consistency.

\subsection{Overall Performance}

\begin{table*}[t]
\centering
\small
\begin{tabular}{lrccccccc}
\toprule
\textbf{Model} & \textbf{Acc.} & \textbf{Prec.} & \textbf{Rec.} & \textbf{F1} & \textbf{$C_{\text{all}}$} & \textbf{$C_{N \leftrightarrow X}$} & \textbf{$C_{O \leftrightarrow OX}$} & \textbf{NLU Acc.} \\
\midrule
Gemini 2.5 Flash & 99.6 & 100.0 & 99.1 & 99.6 & 99.0 & 99.2 & 99.2 & 51.7 \\
GPT-OSS-20B & 99.5 & 100.0 & 99.0 & 99.5 & 96.5 & 97.1 & 98.1 & 51.6 \\
Gemini 2.5 Pro & 99.3 & 100.0 & 98.6 & 99.3 & 98.3 & 98.8 & 98.5 & 51.9 \\
GLM-4.6 & 99.0 & 100.0 & 97.8 & 98.9 & 95.8 & 96.5 & 97.5 & 52.2 \\
Kimi-K2-Instruct & 96.0 & 97.0 & 94.5 & 95.7 & 88.3 & 93.1 & 90.6 & 54.9 \\
DeepSeek V3.1 & 95.8 & 99.6 & 91.6 & 95.4 & 89.0 & 92.1 & 91.7 & 55.1 \\
\midrule
Gemini 2.5 Flash Lite & 88.9 & 89.8 & 86.5 & 88.1 & 71.9 & 82.9 & 77.7 & 57.2 \\
Qwen3-Next 80B A3B Instruct & 79.4 & 73.3 & 88.9 & 80.4 & 69.2 & 81.0 & 76.5 & 46.8 \\
Qwen3-Next 80B A3B Thinking & 72.7 & 99.2 & 42.8 & 59.8 & 76.7 & 81.9 & 85.4 & 64.5 \\
\midrule
Llama 3.3 70B Instruct & 69.8 & 82.1 & 46.7 & 59.5 & 66.2 & 81.0 & 78.3 & 66.3 \\
Gemma 3 27B IT & 68.4 & 61.0 & 93.1 & 73.7 & 69.0 & 82.5 & 86.0 & 43.6 \\
Llama 3.1 8B Instruct & 64.3 & 66.3 & 50.7 & 57.4 & 51.9 & 75.6 & 62.1 & 56.8 \\
Llama 3.2 3B Instruct & 59.2 & 88.1 & 16.2 & 27.4 & 75.0 & 92.1 & 81.7 & 73.7 \\
Llama 3.2 1B Instruct & 51.9 & 49.2 & 41.9 & 45.3 & 57.9 & 76.7 & 73.8 & 60.4 \\
\bottomrule
\multicolumn{9}{l}{\footnotesize All metrics in \%. Acc. = Syntax Accuracy, Prec. = Precision, Rec. = Recall.} \\
\multicolumn{9}{l}{\footnotesize Consistency metrics: $C_{\text{all}}$ (all 4 variants), $C_{N \leftrightarrow X}$ (normal $\leftrightarrow$ nonsense), $C_{O \leftrightarrow OX}$ (order-switched variants).} \\
\end{tabular}
\caption{Comprehensive model performance metrics aggregated across all 12 configurations (4 strategies $\times$ 3 temperatures). Syntax accuracy and NLU accuracy represent dual evaluation frameworks. Models grouped by performance tier.}
\label{tab:overall_performance}
\end{table*}

Performance exhibits a bimodal distribution across the 14 evaluated models (Table~\ref{tab:overall_performance}). Six models achieve above 95\% syntax accuracy, forming a distinct top-tier with robust syllogistic reasoning capability. Gemini 2.5 Flash attains near-perfect performance (99.6\%), deviating from perfect accuracy in fewer than five instances per 1000. At the opposite extreme, five models score below 70\%, with Llama 3.2 1B Instruct performing at 51.9\%. The overall mean syntax accuracy is 81.7\% ($SD$ = 17.1\%), but the 47.7\% gap between top and bottom performers demonstrates that syllogistic reasoning capability depends critically on architectural choices and training methods rather than raw model scale.

The pattern of precision, recall, and F1 scores reveals systematic biases. Qwen3-Next 80B A3B Thinking shows 99.2\% precision but only 42.8\% recall, indicating it labels most syllogisms as \textit{``incorrect''} even when valid. Conversely, Gemma 3 27B IT exhibits 93.1\% recall but only 61.0\% precision, suggesting over-acceptance of conclusions. Top-tier models maintain balanced precision-recall profiles (both $>$97\%), demonstrating genuine discriminative capability.

\subsubsection{Dual Evaluation Framework}

\begin{figure}[t]
\centering
\includegraphics[width=0.9\columnwidth]{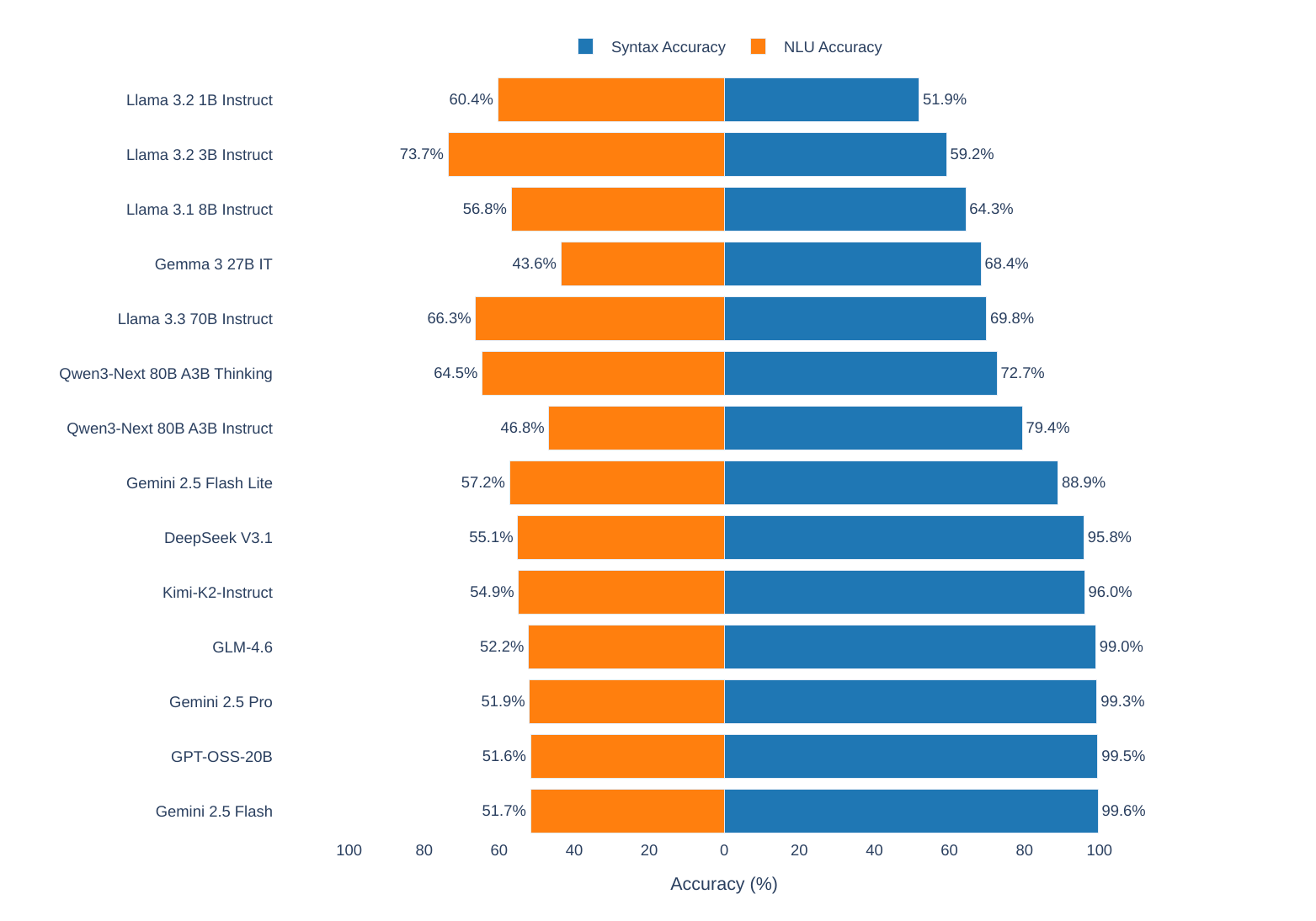}
\caption{Syntactic validity (left) versus natural language understanding believability (right). The 25.50pp gap (syntax: 81.7\%, NLU: 56.2\%) demonstrates that models excel at formal logical reasoning while struggling with semantic plausibility judgments.}
\label{fig:syntax_nlu}
\end{figure}

We evaluated each prediction against both ground truths independently: \textit{syntactic validity} and \textit{NLU believability} (see \S\ref{subsec:dualgroundtruth}). As shown in Figure~\ref{fig:syntax_nlu} and Table~\ref{tab:overall_performance} (final column), syntax accuracy (81.7\%) substantially exceeds NLU accuracy (56.2\%). Top-tier models show large syntax-NLU gaps: Gemini 2.5 Flash (47.9 pp), GPT-OSS-20B (47.9 pp), and Gemini 2.5 Pro (47.4 pp) excel at syntax but perform near chance on NLU evaluation. This pattern emerges because these models correctly judge logical validity independent of content believability. Conversely, three models exhibit negative gaps: Llama 3.2 3B Instruct ($-14.5$ pp), Llama 3.2 1B Instruct ($-8.5$ pp), and Llama 3.3 70B Instruct ($+3.5$ pp shows minimal gap), suggesting that lower-tier models may rely more heavily on semantic plausibility heuristics.

\subsection{Prompting Strategy Effects}

Contrary to expectations, FS prompting yields the lowest mean accuracy (79.1\%), while ZS achieves 82.7\%. A paired $t$-test confirms that FS significantly underperforms ZS ($\Delta = -3.57$ pp, $t_{41} = 2.50$, $p = 0.0165$), with the effect surviving Holm-Bonferroni correction for three comparisons ($p_{\text{adj}} = 0.0495$, Cohen's $d = -0.39$). However, a Friedman test shows no significant overall strategy effect across all four strategies ($\chi^2 = 3.24$, $df = 3$, $p = 0.356$), and Wilcoxon signed-rank tests reveal the effect becomes marginally non-significant after correction ($p = 0.0195$, $p_{\text{adj}} = 0.0584$). Figure~\ref{fig:strategy_heatmap} illustrates the lack of systematic strategy effects across models.

To understand this pattern, we employed McNemar's test at the instance level ($N = 6720$ syllogism evaluations: 14 models $\times$ 3 temperatures $\times$ 160 syllogisms). We find highly significant error redistribution: ZS solves 786 instances that FS fails, while FS solves only 546 that ZS fails ($\chi^2 = 42.88$, $p < 0.0001$). The reconciliation is straightforward: FS prompting changes \textit{which} problems are solved (McNemar test) and produces a consistent directional decline in mean accuracy ($t$-test), but the median effect is less robust (Wilcoxon test). Strategy effects appear model-specific rather than universal.

\subsection{Temperature and Belief Bias Effects}

Temperature ($\tau$) has negligible impact on accuracy when adaptive stopping is employed. A Friedman test confirms no significant temperature effect ($\chi^2 = 3.77$, $df = 2$, $p = 0.152$), with mean accuracy virtually identical across all $\tau$ settings. The adaptive majority-voting mechanism effectively normalizes stochastic variation.


\begin{table}[t]
\centering
\small
\begin{tabular}{lrrr}
\toprule
\textbf{Model} & \textbf{Cong.} & \textbf{Incong.} & \textbf{$\Delta_{\text{bias}}$} \\
\midrule
Llama 3.2 3B Instruct & 82.0 & 35.2 & +46.9 \\
Llama 3.3 70B Instruct & 85.3 & 53.6 & +31.6 \\
Qwen3-Next 80B A3B Thinking & 86.3 & 58.3 & +28.0 \\
Llama 3.2 1B Instruct & 62.0 & 41.2 & +20.8 \\
Llama 3.1 8B Instruct & 70.6 & 57.7 & +12.9 \\
Gemini 2.5 Flash Lite & 95.0 & 82.5 & +12.5 \\
DeepSeek V3.1 & 99.7 & 91.8 & +7.9 \\
Kimi-K2-Instruct & 99.6 & 92.1 & +7.5 \\
GLM-4.6 & 99.4 & 97.5 & +1.9 \\
Gemini 2.5 Pro & 100.0 & 98.6 & +1.4 \\
Gemini 2.5 Flash & 100.0 & 99.2 & +0.9 \\
GPT-OSS-20B & 99.2 & 98.4 & +0.8 \\
\midrule
Qwen3-Next 80B A3B Instruct & 75.5 & 83.4 & $-7.9$ \\
Gemma 3 27B IT & 61.7 & 75.4 & $-13.7$ \\
\bottomrule
\multicolumn{4}{l}{\footnotesize All values in \%. Cong. = Congruent, Incong. = Incongruent.} \\
\end{tabular}
\caption{Belief bias analysis showing accuracy on congruent (logic matches intuition) versus incongruent (logic conflicts with intuition) syllogisms. Sorted by bias magnitude.}
\label{tab:belief_bias}
\end{table}

We observe robust evidence of belief bias across the majority of models (Figure~\ref{fig:belief_bias}, Table~\ref{tab:belief_bias}). Twelve of 14 models exhibit positive belief bias i.e., higher accuracy on congruent problems than on incongruent problems. The mean bias effect is $\Delta_{\text{bias}} = +10.81$ pp ($SD$ = 16.32), statistically significant by paired $t$-test ($t_{13} = 2.47$, $p = 0.0280$, Cohen's $d = 0.66$). 


\subsection{Consistency and Benchmark Correlations}

The consistency metrics in Table~\ref{tab:overall_performance} reveal that high-performing models maintain high consistency across content variants. The correlation between syntax accuracy and overall consistency is very strong (Pearson $r = 0.877$, $p < 0.0001$; Spearman $\rho = 0.890$, $p < 0.0001$), indicating that models achieving high accuracy are substantially more stable across variants.

\begin{figure}[t]
\centering
\includegraphics[width=0.9\columnwidth]{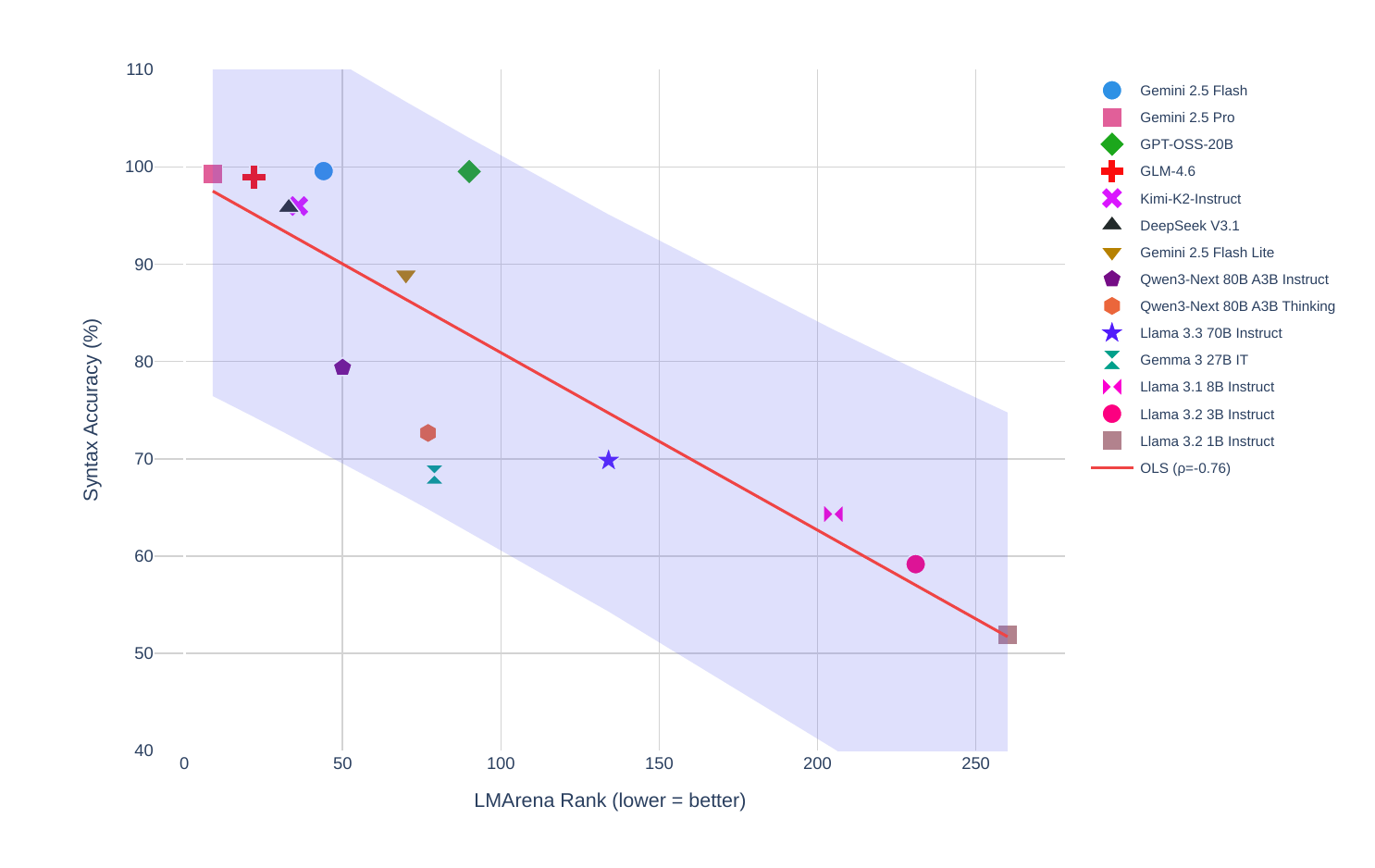}
\caption{Correlation between syllogistic reasoning accuracy and LMArena rankings (Spearman $\rho = -0.825$, $p = 0.0010$, $N = 12$). Lower rank indicates better performance. The strong negative correlation suggests that instruction-following quality predicts formal reasoning capability.}
\label{fig:lmarena}
\end{figure}

To contextualize syllogistic reasoning within the broader LLM evaluation landscape, we computed correlations with LMArena human preference rankings \cite{chiang2024chatbot,zheng2023judging,zheng2024lmsyschatm}. As shown in Figure~\ref{fig:lmarena}, syllogistic reasoning shows a strong negative correlation with LMArena rank (Spearman $\rho = -0.825$, $p = 0.0010$, $N = 12$; lower rank indicates better performance). The negative correlation is the expected as models with higher reasoning accuracy achieve numerically lower (better) LMArena rankings. This suggests that models excelling at instruction following also excel at formal reasoning, likely because both require precise adherence to explicit rules.

\subsection{Statistical Summary}

\begin{table*}[t]
\centering
\small
\begin{tabular}{llccccl}
\toprule
\textbf{Analysis} & \textbf{Test} & \textbf{Statistic} & \textbf{df} & \textbf{$p$-value} & \textbf{Effect} & \textbf{Result} \\
\midrule
\multicolumn{7}{l}{\textit{Main Effects}} \\
Strategy effect (overall) & Friedman $\chi^2$ & 3.24 & 3 & 0.356 & --- & No effect \\
ZS vs FS & Paired $t$ & 2.50 & 41 & 0.0165$^*$ & $d = -0.39$ & Significant \\
ZS vs FS (Holm) & Paired $t$ & 2.50 & 41 & 0.0495$^*$ & $d = -0.39$ & \textbf{Survives correction} \\
Temperature effect & Friedman $\chi^2$ & 3.77 & 2 & 0.152 & --- & No effect \\
Belief bias (Cong. $>$ Incong.) & Paired $t$ & 2.47 & 13 & 0.0280$^*$ & $d = 0.66$ & \textbf{Confirmed} \\
\midrule
\multicolumn{7}{l}{\textit{McNemar Tests (Instance-level, $N = 6720$)}} \\
ZS vs FS & McNemar $\chi^2$ & 42.88 & 1 & $<$0.0001$^{***}$ & 786 vs 546 & \textbf{Error redistribution} \\
ZS vs OS & McNemar $\chi^2$ & 1.70 & 1 & 0.192 & 317 vs 284 & No redistribution \\
ZS vs ZS CoT & McNemar $\chi^2$ & 0.26 & 1 & 0.612 & 389 vs 374 & No redistribution \\
\midrule
\multicolumn{7}{l}{\textit{Key Correlations ($N = 14$ models)}} \\
Syntax Acc. $\times$ Overall Consistency & Spearman $\rho$ & 0.890 & --- & $<$0.0001$^{***}$ & Very strong & Positive \\
Syntax Acc. $\times$ $C_{N \leftrightarrow X}$ & Spearman $\rho$ & 0.846 & --- & 0.0001$^{***}$ & Very strong & Positive \\
Syntax Acc. $\times$ $C_{O \leftrightarrow OX}$ & Spearman $\rho$ & 0.837 & --- & 0.0002$^{***}$ & Very strong & Positive \\
Syntax Prec. $\times$ Syntax Rec. & Spearman $\rho$ & 0.691 & --- & 0.0062$^{**}$ & Strong & Positive \\
Syntax Acc. $\times$ NLU Acc. & Spearman $\rho$ & $-0.543$ & --- & 0.0449$^*$ & Moderate & \textbf{Negative} \\
Syntax Acc. $\times$ Bias Effect & Spearman $\rho$ & $-0.565$ & --- & 0.0353$^*$ & Moderate & \textbf{Negative} \\
\midrule
\multicolumn{7}{l}{\textit{Benchmark Correlation}} \\
LMArena rank (lower = better) & Spearman $\rho$ & $-0.825$ & --- & 0.0010$^{***}$ & Very strong & \textbf{Predicts reasoning} \\
\bottomrule
\multicolumn{7}{l}{\footnotesize $^*p < 0.05$, $^{**}p < 0.01$, $^{***}p < 0.001$. Holm-Bonferroni correction applied to strategy comparisons.} \\
\multicolumn{7}{l}{\footnotesize McNemar instances: ``786 vs 546'' = ZS correct \& FS wrong vs FS correct \& ZS wrong.} \\
\multicolumn{7}{l}{\footnotesize Bias correlation: Negative $\rho$ means higher accuracy correlates with smaller bias magnitude (closer to zero).} \\
\end{tabular}
\caption{Comprehensive statistical summary of all hypothesis tests and correlations for 14 models. Strategy comparisons use Holm-Bonferroni correction. McNemar test operates at instance-level (6,720 syllogism evaluations per comparison).}
\label{tab:statistical_summary}
\end{table*}

Table~\ref{tab:statistical_summary} consolidates all key statistical findings. The FS underperformance survives Holm-Bonferroni correction ($p_{adj} = 0.0495$), while the McNemar test reveals significant error redistribution at the instance level. The reconciliation between significant $t$-test and marginally non-significant Wilcoxon test ($p_{raw} = 0.0195$, $p_{adj} = 0.0584$) reveals that FS produces a consistent mean decline but less robust median effect.

The correlation between syntax accuracy and belief bias magnitude shows a moderate negative relationship (Spearman $\rho = -0.565$, $p = 0.0353$). Since bias effect is defined as $Acc_{congruent} - Acc_{incongruent}$, this negative correlation indicates that higher performing models exhibit smaller bias magnitudes. It further provides evidence that higher reasoning ability reduces reliance on content based heuristics.

The very strong correlations between syntax accuracy and all three consistency metrics ($\rho = 0.890$, 0.846, and 0.837, all $p < 0.001$) confirm that models achieving high accuracy are substantially more stable across content and order variations. The moderate negative correlation between syntax and NLU accuracy (Spearman $\rho = -0.543$, $p = 0.0449$) indicates that models optimized for logical structure may diverge from intuitive believability judgments.

\section{Discussion}
\label{sec:discussion}

In this study, we analyzed 40 instances of syllogism and their variations, resulting in a total of 160 data points tested against 14 different large language models. Our results demonstrate a striking pattern: top-tier models achieve near-perfect syntactic accuracy (99.6\%) while performing at chance levels on natural language understanding (52\%). This behavior, excelling at formal logic while struggling with semantic plausibility, contrasts sharply with human reasoning, where belief bias typically dominates logical analysis.

The majority of models exhibit significant belief bias, performing better when logic aligns with intuition (mean effect: +10.81 pp, $p=0.028$). However, this bias decreases systematically with improved reasoning capability ($\rho = -0.565$, $p = 0.035$), suggesting that higher-performing models increasingly prioritize formal rules over semantic heuristics. Architectural and training choices prove more consequential than raw parameter count by substantial margins. Counterintuitively, few-shot prompting degraded performance compared to zero-shot, suggesting demonstration examples may introduce noise in formal reasoning tasks. The strong correlation between instruction following quality (LMArena, $\rho = -0.825$) and reasoning accuracy indicates that precise rule adherence underlies both capabilities.

These findings suggest that most models exhibit a preference for symbolic reasoning and inferences rather than adhering to the natural language path of reasoning characteristic of human cognition. While this result may appear promising from a purely logical perspective, it raises important questions about the alignment between LLM reasoning and human cognitive processes. These models were trained on extensive natural language data, yet the top performers appear to function more like formal logic engines than human-like reasoners susceptible to known natural language biases.


\section{Limitations}
\label{sec:limitations}

Our evaluation focuses primarily on categorical syllogisms, a narrow subset of logical reasoning that may not generalize to more complex structures with nested quantifiers or modal operators. The dual ground truth framework, while enabling systematic measurement, necessarily simplifies the dynamic interaction between logic and natural language that humans navigate simultaneously in real reasoning contexts.

The scope of our study includes only 14 models, representing a snapshot of the current LLM landscape but not exhaustive coverage of all available systems. Our prompting strategies, while covering major paradigms (zero-shot, one-shot, few-shot, chain-of-thought), constitute a limited exploration of the vast prompt engineering space. Additionally, our consistency metrics measure stability across content and order variations but do not assess robustness to adversarial perturbations or systematically manipulated distractors.

The belief bias metric, while grounded in cognitive psychology literature, captures only one dimension of the complex relationship between real world beliefs and logical reasoning. Future work should incorporate additional measures such as response time analysis, confidence calibration, and fine-grained error taxonomies to provide a more comprehensive understanding of LLM reasoning processes.

\section{Future Work}
\label{sec:futurework}

Several promising directions emerge from this work. Extending evaluation to richer logical systems such as modal logics, transitive closure logics, to test whether the observed patterns generalize beyond categorical syllogisms. Particular interest lies in logical systems with simple formal syntax but complex natural language semantics, which would further stress the formal logic-natural language divide that we observed.

Complementing these empirical extensions, mechanistic interpretability studies could reveal whether models learn explicit logical rules, statistical approximations, or hybrid representations. This would clarify the computational basis of the near-perfect syntactic performance we documented in top-tier models. Related to this, the causal relationship between reasoning capability and bias resistance remains an open question: does logical training reduce bias, or does reduced bias enable better reasoning? Controlled fine-tuning experiments could disentangle these possibilities.

Our finding that few-shot prompting degraded performance challenges conventional wisdom and warrants systematic exploration of when and why demonstration examples help versus hinder reasoning. Such investigation would inform more effective prompting strategies for logical reasoning tasks.

More broadly, our results raise a fundamental tension i.e., are we building human like reasoners or formal logic engines? This question has implications not only for model development but also for appropriate deployment contexts and expectations for LLM behavior in reasoning-intensive applications. We intend to continue this line of inquiry across other logical reasoning tasks to better understand the trajectory of the cognitive capabilities of LLM.


\section*{Acknowledgements}
We thank the Indo-French Centre for the Promotion of Advanced Research (IFCPAR/CEFIPRA) for their support. This work was supported through project number CSRP-6702-2.


\bibliography{aaai2026}



\end{document}